\documentclass[runningheads]{llncs}
\usepackage{graphicx}
\usepackage{amsmath,amssymb} 
\usepackage{url}
\usepackage{cite}
\usepackage{color}
\usepackage{epsfig}
\usepackage{graphicx}
\usepackage{wrapfig}
\usepackage{amsmath}
\usepackage{amssymb}
\usepackage{multirow}
\usepackage{booktabs}
\usepackage{algorithmic}
\usepackage{arydshln}
\usepackage{threeparttable}
\usepackage{colortbl}
\usepackage{enumitem}
\usepackage{siunitx}
\usepackage{bm}
\usepackage{array}
\usepackage{colortbl}
\usepackage{dblfloatfix}
\usepackage{marvosym}
\usepackage{lipsum}
\definecolor{mygray}{gray}{.9}
\definecolor{ggray}{RGB}{127,127,127}
\definecolor{reda}{RGB}{192,0,0}
\definecolor{redb}{RGB}{217,148,143}
\definecolor{myyellow}{RGB}{190,144,0}
\definecolor{mygreen}{RGB}{80,100,40}
\definecolor{myblue}{RGB}{30,90,100}
\newcommand{\tabincell}[2]{\begin{tabular}{@{}#1@{}}#2\end{tabular}}

\newcommand{\ie}{\textit{i}.\textit{e}.}
\newcommand{\eg}{\textit{e}.\textit{g}.}

\usepackage[pagebackref=true,breaklinks=true,letterpaper=true,colorlinks,bookmarks=false]{hyperref}
\usepackage[width=122mm,left=12mm,paperwidth=146mm,height=193mm,top=12mm,paperheight=217mm]{geometry}

\makeatletter
\newcommand{\thickhline}{%
	\noalign {\ifnum 0=`}\fi \hrule height 1pt
	\futurelet \reserved@a \@xhline
}
\makeatother

\begin{document}

\pagestyle{headings}
\mainmatter
\def\ECCVSubNumber{1062}  

\title{Video Object Segmentation with\\ Episodic Graph Memory Networks} 
\titlerunning{Video Object Segmentation with Episodic Graph Memory Networks}
\authorrunning{X. Lu, W. Wang, M. Danelljan, T. Zhou, J. Shen,  L. Van Gool}

\author{Xiankai Lu$^{1}$~\!,\hspace{1pt} \Letter Wenguan Wang$^{2}$~\!, \hspace{1pt} Martin Danelljan$^{2}$\\
Tianfei Zhou$^{1}$~\!, Jianbing Shen$^{1}$~\!,  and Luc Van Gool$^{2}$}

\institute{{$^1$} Inception Institute of Artificial Intelligence \hspace{2pt} {$^2$} ETH Zurich\\
\{carrierlxk, wenguanwang.ai\}@gmail.com\\
\url{https://github.com/carrierlxk/GraphMemVOS}}
\maketitle
\begin{abstract}
How to make a segmentation model efficiently adapt to a specific video as well as online target appearance variations is a fundamental issue in the field of video object segmentation. In this work, a graph memory network is developed to address the novel idea of ``learning to update the segmentation model''. Specifically, we exploit an episodic memory network, organized as a fully connected graph, to store frames as nodes and capture cross-frame correlations by edges. Further, learnable controllers are embedded to ease memory reading and writing, as well as maintain a fixed memory scale. The structured, external memory design enables our model to comprehensively mine and quickly store new knowledge, even with limited visual information, and the differentiable memory controllers slowly learn an abstract method for storing useful representations in the memory and how to later use these representations for prediction, via gradient descent. 
In addition, the proposed graph memory network yields a neat yet principled framework, which can generalize well to both one-shot and zero-shot video object segmentation tasks.
Extensive experiments on four challenging benchmark datasets verify that our graph memory network is able to facilitate the adaptation of the segmentation network for case-by-case video object segmentation.
\end{abstract}
\newcommand\blfootnote[1]{%
\begingroup
\renewcommand\thefootnote{}\footnote{#1}%
\addtocounter{footnote}{-1}%
\endgroup
}
\section{Introduction}
Video object segmentation (VOS)\blfootnote{\Letter~Corresponding author: \textit{Wenguan Wang}.}, as a core task in computer vision, aims to predict the target object in a video at the pixel level. 
Typically, according to whether or not annotations are provided for the first frame during testing, VOS can be categorized into one-shot video object segmentation (O-VOS)\!~\cite{Cae+17,ventura2019rvos} and zero-shot video object segmentation (Z-VOS)\!~\cite{wang2019zero}. Provided with only first-frame annotations, O-VOS is to identify and segment the labeled  object instances in the rest of the video\!~\cite{perazzi2017learning,DBLP:conf/iccv/YoonRKLSK17,cheng2018fast,Hu_2018_ECCV,maninis2018video,johnander2019generative}; whereas Z-VOS targets at automatically inferring the primary objects without any test-time indications\!~\cite{DBLP:conf/cvpr/TokmakovAS17,DBLP:conf/iccv/TokmakovAS17}.

O-VOS is a challenging task because there are no further assumptions regarding to the specific target object and the application scenes often contain similar distractor objects. To tackle these challenges, earlier methods typically perform network finetuning over each annotated object\!~\cite{Cae+17,voigtlaender2017online}. Though effective, this is quite time-consuming. Current popular solutions\!~\cite{voigtlaender2019feelvos,wang2019fast,Oh_2019_ICCV,Zeng_2019_ICCV,Duarte_2019_ICCV} are instead built upon an efficient \textit{matching} based framework; they formulate the task as a differentiable matching procedure between the support set (\ie, the first labeled frame or prior segmented frames) and query set (\ie, current frame). Thus they can directly assign labels to the query frame, according to the pixel-wise similarity to the annotated first frame and/or previous processed frames.

Although omitting first-frame finetune and improving the performance to some extent, matching based O-VOS methods still suffer from several limitations. First, they typically learn a generic matching network and then apply it to test videos directly, failing to make full use of first-frame target-specific information. As a result, they cannot efficiently adapt to the input video. Second, as the segmentation targets may undergo appearance variation (\ie, fast motion, occlusion), it is meaningful to perform online model updating. Third, matching based methods only modeling pair-relations between the query and each support frame, neglecting the rich context within the support set. 

To address these issues, we take inspiration from the recent development of memory-augmented networks for few-shot learning\!~\cite{weston2014memory,santoro2016meta} and develop a graph memory network to online adapt the segmentation model to a specific target in one single feed-forward pass. Specifically, by regarding O-VOS as episodic memory reasoning, our approach equips with the ability to slowly learn high-level knowledge for extracting useful representations from the offline training data, and the ability to rapidly fuse the unseen information from the first-frame annotation in the test video, via an external memory module. In this way, our model can internally modulate the output by learning to rapidly cache representation in the memory stores. During the segmentation, to maintain the variations of the object appearance, we perform memory updating by  storing and recalling target information from the external memory.  Therefore, we can implement the online model updating easily without extensive parameter optimization. In addition, the memory module, built upon the end-to-end memory network\!~\cite{sukhbaatar2015end}, is endowed with a graph structure to better mine the relations among memory cells.

The proposed graph memory network is neat and fast. For memory updating, instead of some prior matching based O-VOS models\!~\cite{Oh_2019_ICCV} inserting a new element into newly allocated position,  our model performs message passing on the fixed-size graph memory without increasing memory consumption. Our model provides a principled framework; it generalizes Z-VOS task well, in which mainstream methods also lack the adaptation capability. As far as we know, this work represents the first effort that addresses both O-VOS and Z-VOS in a unified network design. Experiments on representative O-VOS datasets show the proposed method performs favorably against state-of-the-arts. Without bells and whistles, it also outperforms other competitors on Z-VOS datasets. These promising results demonstrate the efficacy and generalizability of our graph memory network.

\section{Related Work}
\noindent\textbf{One-shot Video Object Segmentation (O-VOS)} is to track the first-frame annotation to subsequent frames at pixel level. Traditional methods usually formulate this task as a \textit{label propagation} process\!~\cite{wen2015jots,DBLP:conf/cvpr/BadrinarayananGC10,shankar2015video,perazzi2015fully,wang2018semi}. With the renaissance of the connectionism, deep learning based solutions become the domaniant\!~\cite{perazzi2017learning,perazzi2015fully,maninis2018video,Duarte_2019_ICCV} in the field of O-VOS.  Among them are three representative strategies. One is the \textit{segmentation-by-detection} scheme\!~\cite{Cae+17,Chen_2018_CVPR,luiten2018premvos,ci2018video} that learns a video-specific representation about the first-frame annotated objects and then performs pixel-wise detection in the rest frames. Another one is the \textit{propagation}  based pipeline\!~\cite{ci2018video,xiao2018monet,DBLP:conf/iccv/TokmakovAS17}, which propagates  segmented masks to fit objects in the upcoming frames.  The third, \ie, the more advanced, is the \textit{matching} based strategy\!~\cite{DBLP:conf/iccv/YoonRKLSK17,cheng2018fast,Hu_2018_ECCV,johnander2019generative,voigtlaender2019feelvos,wang2019fast,Oh_2019_ICCV,Zeng_2019_ICCV,Duarte_2019_ICCV,lu2020learning} which usually trains a prototypical Siamese matching network to find the most matching pixel (or embedding in the feature space) between the first frame (or a segmented frame) and the query frame, and then achieves label assignment accordingly. 
Some matching-based methods employ internal memory (\eg, ConvLSTM\!~\cite{ventura2019rvos,xu2018youtube-eccv}) or external memory (\eg, \cite{Oh_2019_ICCV,ventura2019rvos}) to implicitly or explicitly store previously computed segmentation information for facilitating learning the evolution of objects over time. However, our utilization of memory differs from these methods substantially: \textbf{i)} we employ an external memory with learnable read-write controller to rapidly encode new video information for quick segmentation network updating; \textbf{ii)} compared to vanilla memory network\!~\cite{sukhbaatar2015end,miller2016key}, our graph memory network stores memory content in a structured manner that explicitly captures context in cells; \textbf{iii)} instead of writing new input to a newly located position\!~\cite{Oh_2019_ICCV}, our memory is dynamically updated by iterative cell state renewing without increasing the memory size.   

\noindent\textbf{Zero-shot Video Object Segmentation (Z-VOS)} aims to segment primary objects in unconstrained videos. This task has been widely studied over several decades which also called unsupervised video object segmentation\!~\cite{lee2011key,zhang2013,perazzi2015fully,DBLP:conf/cvpr/KohK17}. Traditional methods usually leverage motion\!~\cite{DBLP:conf/iccv/OchsB11,DBLP:conf/iccv/KeuperAB15} or saliency cues\!~\cite{DBLP:conf/bmvc/FaktorI14,wang2017saliency} to obtain a heuristic representation for inferring the primary objects. Recent methods were built upon fully convolutional networks. Early methods explored two-stream architectures\!~\cite{jain2017fusionseg,cheng2017segflow,siam2018video,zhou2020motion} or variants of recurrent neural networks\!~\cite{DBLP:conf/iccv/TokmakovAS17,wang2019learning,Song_2018_ECCV}. Recent ones address comprehensive foreground reasoning from a global view by non-local structures\!~\cite{lu2019see,wang2019zero}. In this work, rather than these methods learning a universal video foreground object representation and hoping it could generalize  well to all unseen scenarios, our episodic memory design allows target-adaption on-the-fly by learning to update the segmentation network.


\noindent\textbf{Learning to Update in VOS.} 
To learn more video-specific information, a direct way is to perform iterative network finetune in the first frame\!~\cite{Cae+17,voigtlaender2017online}. Some recent efforts instead applied meta-learning for model updating\!~\cite{yang2018efficient,xiao2019online,behl2018meta}, whose basic ideas are similar: learning segmentation model parameters on training videos. This paper, in contrast, uses a graph memory network with learnable, lightweight controllers to assimilate a new video. 
Thus the model can quickly adapt to unseen scenes and appearance variants  through the memory node (cell) updating rather than sensitive and expensive network parameter generation\!~\cite{yang2018efficient,xiao2019online}.


\section{Proposed Algorithm}
\label{sec:method}

\subsection{Preliminary: Episodic Memory Networks}
\label{sec:E2EMemory}
Memory networks augment neural networks with an external memory component\!~\cite{weston2014memory,sukhbaatar2015end,miller2016key}, which allow the network to explicitly access the past experiences. They have been shown effective in few-shot learning\!~\cite{santoro2016meta,xie2019attentive,xie2020region} and object tracking\!~\cite{Yang2018}. Recently, episodic external memory networks have been explored to solve reasoning issues in visual question answering and visual dialog\!~\cite{sukhbaatar2015end,kumar2016ask,miller2016key}. The basic idea is to retrieve the information required to answer the question from the memory with trainable read and write operators.  Given a collection of input representations, the episodic memory module chooses which parts of the inputs to focus on through the neural attention. It then produces a ``memory summarization'' representation taking into account the query as well as the stored memory. Each iteration in the episode provides the memory module with newly relevant information about the input. As a result, the memory module has the ability to retrieve new information in each iteration and obtain a new representation about the input. 

\subsection{Learning to Update}
\label{sec:learn-to-update}
\begin{figure*}[tbp]
	\centering
	\includegraphics[width=.99\textwidth]{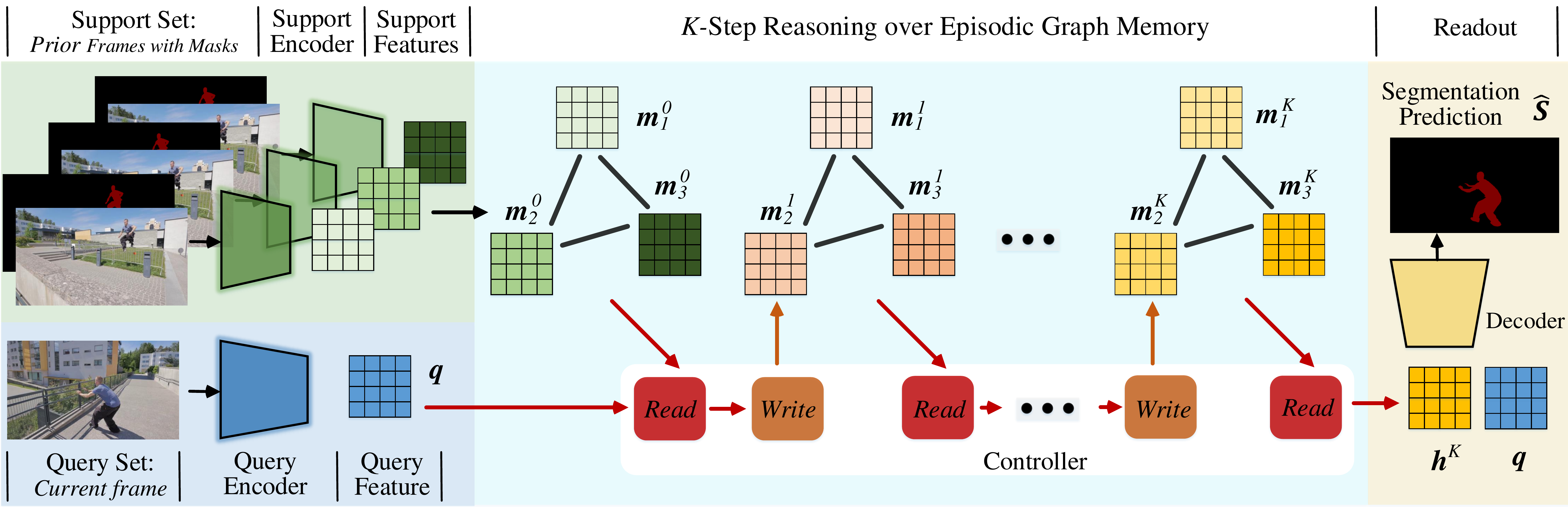}
	\caption{\textbf{$_{\!}$Illustration$_{\!}$ of$_{\!}$ our$_{\!}$ graph$_{\!}$ memory$_{\!}$} based$_{\!}$ O-VOS$_{\!}$ method. Previous frames are fed together with the pre-defined or self-segmented masks to the support encoder to initialize graph memory nodes $\{\bm{m}^0_i\}_i$.  Current frame is fed into the query encoder to output the query embedding $\bm{q}$. The graph memory interacts with $\bm{q}$ under several episodic reasoning (with learnable read and write controllers) to mine$_{\!}$ support$_{\!}$ context$_{\!}$ and$_{\!}$ generate$_{\!}$ video specific features. After $K$-step episodic reasoning, the decoder~predicts segmentation mask $\bm{\hat{S}}$, based on the episodic feature $\bm{h}^{K\!\!}$ and query embedding~$\bm{q}$.\!\!
}
	\label{fig:framework}
\end{figure*}

In the context of O-VOS\!~\cite{Cae+17}, the goal is to learn from the annotated objects in the first frame (\textit{support set}) and predict them in the subsequent frames (\textit{query set}).   To this end, traditional methods usually finetune the network and perform online learning for each specific video. In contrast, we construct an episodic memory based learner on variety of tasks (videos), sampled from the distribution of training tasks\!~\cite{behl2018meta}, such that the learned model performs well on new unseen tasks (test videos). Thus we tackle O-VOS as a \textbf{``learning to update''} segmentation network procedure\!~\cite{rakelly2019metalearning}: \textbf{i)} extracting a task representation from the one-shot support set, and \textbf{ii)} updating the segmentation network for the query given the task representation. 
As shown in {Fig.~\ref{fig:framework}}, we enhance an episodic memory network with graph structure (\ie, graph memory network) to: \textbf{i)} instantly adapt the segmentation network to a specific object, rather than performing lots of finetuning iterations; and \textbf{ii)} make full use of context within video sequences. As a result, our graph memory network has two abilities: learn to adjust the segmentation network from one-shot support set during the model initialization phase and learn to take advantage of segmented frames to update the segmentation during frame processing phase. Our model thus can make efficient case-by-case adaption and online updating within one feed-forward process. 

\subsection{Graph Memory Network}
\label{sec:graph-mmeory}

The graph memory network consists of an external graph memory and learnable controllers for memory operating. The memory provides short-term storage for new knowledge encoding and its graph structure allows to fully explore the context. The controllers interact with the graph memory through a number of read and write operations and they are capable of long-term storage via slow updates of the weights. Through the controllers, our model learns a general strategy for the types of representations it should place into the
memory and how it should later use these representations for segmentation predictions.

The core idea of our graph memory network is to perform $K$ steps of episodic reasoning to efficiently mine the structures in the memory and better capture target-specific information. Specifically, the memory is organized as a size-fixed, fully connected graph $\mathcal{G}\!=\!(\mathcal{M},\mathcal{E})$, where node $m_i\!\in\!\mathcal{M}$ denotes $i^{th}$ memory cell, and edge $e_{i,j}\!=\!(m_i,m_j)\!\in\!\mathcal{E}$ indicates the relation between cell $m_i$ and $m_j$.

Given a query frame, the support set is considered as the combination of the first annotated frame and previously segmented frames. The graph memory is initialized from $N(=_{\!}|\mathcal{M}|)$ frames,  sampled from the support set. For each memory node $m_i$, its initial embedding $\bm{m}^0_i\!\in\!\mathbb{R}^{W\!\times\! H\! \times\!C}$ is generated by applying a fully convolutional memory encoder to the corresponding support frame to capture both the spatial visual feature as well as segmentation mask information.

\noindent\textbf{Graph Memory Reading.} A fully convolutional query encoder is also applied to the query frame to extract the visual feature  $\bm{q}\!\in\!\mathbb{R}^{W\!\times\!H\!\times\!C}$. A learnable \textit{read controller} first takes $\bm{q}$ as input and generates its initial state $\bm{h}^0$:
\begin{equation}
	\bm{h}^0 = f_{\text{P}}(\bm{q})\in\mathbb{R}^{W\times H \times C},
	\label{equ:query-encode}
\end{equation}
where $f_{\text{P}}(\cdot)$ indicates a projection function.  

At each episodic reasoning step $k\!\in\!\{1,...,K\}$, the read controller interacts with the external graph memory by reading the content. Following the key-value retrieval mechanism in\!~\cite{sukhbaatar2015end,miller2016key,kumar2016ask}, we first compute the similarity between the query and each memory node $m_i$:
\begin{equation}
	s^k_i = \frac{\bm{h}^{k-1}\cdot \bm{m}^{k-1}_i}{\|\bm{h}^{k-1}\| \| \bm{m}^{k-1}_i\|}\in[-1,1].
	\label{eq:distance}
\end{equation}
Next, we compute the read weight $w^k_i$ by a softmax normalization function:
 \begin{equation}
 	w^k_i=\exp(s^k_i)\Big/\sum\nolimits_j{\exp(s^k_j)}\in[0,1].
\label{eq:read-weight}
 \end{equation}
Considering some nodes are noisy due to the underlying camera
shift or out-of-view, $w^{k\!}_i$ measures the
confidence of memory cell $m_i$. Then  the memory summarization $\bm{m}^{k\!}$ is retrieved using this weight to linearly combine the memory~cell:\!\!
\begin{equation}
\bm{m}^k = \sum\nolimits_i w^k_i \bm{m}^{k-1}_i\in\mathbb{R}^{W\times H \times C}.
\label{eq:retrival}
\end{equation}

\begin{figure*}[tbp]
	\centering
	\includegraphics[width=.99\textwidth]{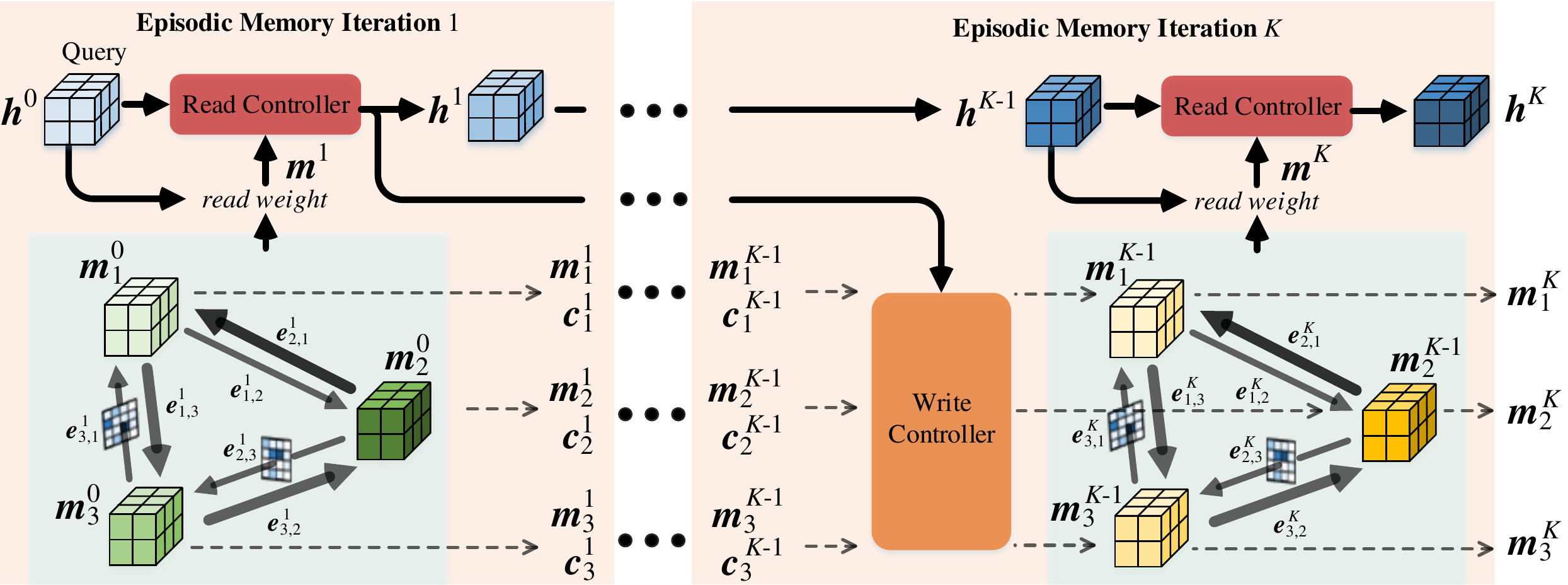}
	\caption{\textbf{Illustration of iterative reasoning} over the episodic graph memory.}
	\label{fig:graph-memory}
\end{figure*}

Through Eqs.~(\ref{eq:distance}-\ref{eq:retrival}), the memory module retrieves the memory cell most similar to $\bm{h}^k$ to obtain the memory summarization $\bm{m}^k$.
Once reading the memory summarization, the read controller updates its state as follows:
\begin{equation}
\begin{aligned}
	\bm{\widetilde{h}}^k &=\bm{W}_r^h*\bm{h}^{k-1}+\bm{U}^h_r*\bm{m}^k~~~~~\in\mathbb{R}^{W\times H \times C}, \\
    \bm{{a}}^k_r &= \sigma(\bm{W}^a_r*\bm{h}^{k\!-\!1}+\bm{U}^a_r*\bm{m}^k)~~\in[0,1]^{W\times H \times C},\\
	\bm{h}^k &= \bm{{a}}_r^k\circ\bm{\widetilde{h}}^k + (\bm{1}- \bm{{a}}_r^k)\circ\bm{h}^{k-1}\!~~\in\mathbb{R}^{W\times H \times C},
	\label{equ:controller-update}
\end{aligned}
\end{equation}
where $\bm{W}$s and $\bm{U}$s are convolution kernels and $\sigma$ indicates the \textit{sigmoid} activation function. `$*$' and `$\circ$' represent the convolution operation and Hadamard product, respectively.
The  update gate $\bm{{a}}^k$ controls how much previous hidden state $\bm{h}^{k-1\!}$ to be kept. In this way, the hidden state of the controller encodes both the graph memory and query representations, which are necessary to produce an output.

\noindent\textbf{Episodic Graph Memory Updating.}
After each pass through the memory summarization, we need to update the episodic graph memory with the new query input. At each step $k$, a learnable memory \textit{write controller} updates each memory cell (\ie, graph node) $m_i$ by considering its previous state $\bm{m}^{k-1}_i$, current content from the read controller $\bm{h}^k$, and the states from other cells $\{\bm{m}^{k-1}_j\}_{j\neq i}$. Specifically, following\!~\cite{wang2019zero}, we first formulate the relation $\bm{e}^k_{i,j}$ from $m_j$ to $m_i$ as the inner-product similarity over their feature matrices:
\begin{equation}
	\begin{aligned}
		\bm{e}_{i,j}^k \!=\! {\bm{m}^{\!k-1}_i}\!~\bm{W}_e\!~{\bm{m}_j^{\!k-1\top}} \in \mathbb{R}^{(W\!H) \times (W\!H)},
	\end{aligned}
	\label{edge}
\end{equation}
where $\bm{W}_{\!e\!\!} \in\!\mathbb{R}^{C\times C\!}$ indicates a learnable weight matrix, and $\bm{m}^{k-1\!}_{i\!\!}\!\in_{\!}\!\mathbb{R}^{{(W\!H) \times C}\!\!}$ and $\bm{m}^{k-1}_{j\!\!}\!\in_{\!}\! \mathbb{R}^{{(W\!H)}\times C\!}$  are$_{\!}$ flattened$_{\!}$ into$_{\!}$ matrix$_{\!}$ representations.
$\bm{e}_{i,j}^k$ stores similarity scores corresponding to all pairs of positions in $\bm{m}_i$ and $\bm{m}_j$.

Then, for $m_i$, we compute the summarized information $\bm{c}^k_i$ from other cells, weighted by their normalized inner-product similarities:
\begin{equation}
\begin{aligned}
   \bm{c}^k_i = \sum\nolimits_{j\neq i}\text{softmax}(\bm{e}^k_{i,j}) \bm{m}^{k\!-\!1}_j \in\mathbb{R}^{W\times H \times C},
   \end{aligned}
	\label{equ:edge}
\end{equation}
where softmax($\cdot$) normalizes each row of the input.

After aggregating the information from neighbors, the memory write controller updates the state of $m_i$ as:
\begin{equation}
\begin{aligned}
	\bm{\widetilde{m}}^k_i &=\bm{W}^m_u*\bm{h}^{k}+\bm{U}^m_u*\bm{m}^{k-1}_i+\bm{V}^m_u*\bm{c}^{k}_i~\!~\in\mathbb{R}^{W\times H \times C}, \\
    \bm{{a}}^k_u &= \sigma(\bm{W}^a_u*\bm{h}^{k}+\bm{U}^a_u*\bm{m}^{k-1}_i+\bm{V}^a_u*\bm{c}^{k}_i)\in[0,1]^{W\times H \times C},\\
	\bm{m}^k_i &= \bm{{a}}^k_u\circ\bm{\widetilde{m}}^k + (\bm{1}- \bm{{a}}^k_u)\circ\bm{m}^{k-1}~~~~~~~~\!~~~~\in\mathbb{R}^{W\times H \times C}.
\label{equ:mem-updating}
\end{aligned}
\end{equation}
The graph memory updating allows each memory cell to embed the neighbor information into its representation so as to fully explore the context in the support set. Moreover, by iteratively reasoning over the graph structure, each memory cells encode the new query information and yield progressively improved representations. Compared with traditional memory network\!~\cite{weston2014memory}, the proposed graph memory network brings two advantages: \textbf{i)} the memory writing operation is fused into the memory updating procedure without increasing the memory size, and \textbf{ii)} avoiding designing complex memory writing strategies\!~\cite{weston2014memory,kumar2016ask,santoro2016meta}. {Fig.~\ref{fig:graph-memory}} shows a detailed diagram of memory reading and updating.

\noindent\textbf{Final Segmentation Readout.}
After $K$ steps of the episodic memory updating, we leverage the final state $\bm{h}^{K\!}$ from the memory read controller to support the prediction on the query:
\begin{equation}
\bm{\hat{S}} = f_{\text{R}}(\bm{h}^{\!K\!}, \bm{q}) \in [0,1]^{W\times H \times2},
\label{equ:readout}
\end{equation}
where the readout function $f_{\text{R}}(\cdot)$ gives the final segmentation probability maps.

\subsection{Full Network Architecture}
\label{sec:architecture}

\noindent\textbf{Network Configuration.}  Our whole model is end-to-end fully convolutional. Both the query encoder and memory encoder have the same network architecture,  except for the inputs. The query encoder takes an RGB query frame as input, while, for the memory encoder, {{input is an RGB support frame concatenated with the one-channel softmax object mask and one-hot label map~\cite{Oh_2019_ICCV}.}} For the graph memory, the \textit{read controller} (Eq.\!~(\ref{equ:controller-update})) and \textit{write controller} (Eq.\!~(\ref{equ:mem-updating})) are all implemented using ConvGRU\!~\cite{ballas2015delving}, with $1\!\times\!1$ convolutional kernels. The project function $f_{\text{P}}$ (Eq.\!~(\ref{equ:query-encode})) is also realized with $1\!\times\!1$ convolutional layer. Similar to\!~\cite{wug2018fast}, the readout function $f_{\text{R}}$ (Eq.\!~(\ref{equ:readout})) is implemented with a decoder network, which consists of four blocks with skip connections to the corresponding ResNet50\!~\cite{he2016deep} blocks. The kernel size of each convolution layer in the decoder is set as $3\!\times\!3$, excepting the last $1\!\times\!1$ convolution layer. The final 2-channel segmentation prediction is obtained by a softmax operation. The query and memory encoders are implemented as the four convolution blocks of ResNet50, initialized by the weights pretrained on ImageNet. The other layers are randomly initialized. Considering memory encoder takes binary mask and instance label maps as input, extra $1\!\times\!1$ convolutional layers are used for encoding these inputs. The resulting features are added with RGB features at the first blocks of ResNet50.

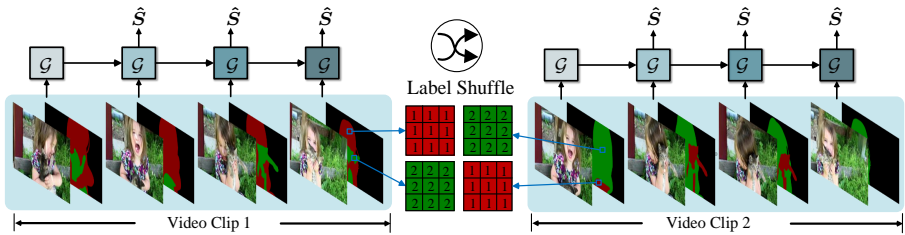
\begin{figure*}[t]
	\centering
	\includegraphics[width=.99\textwidth]{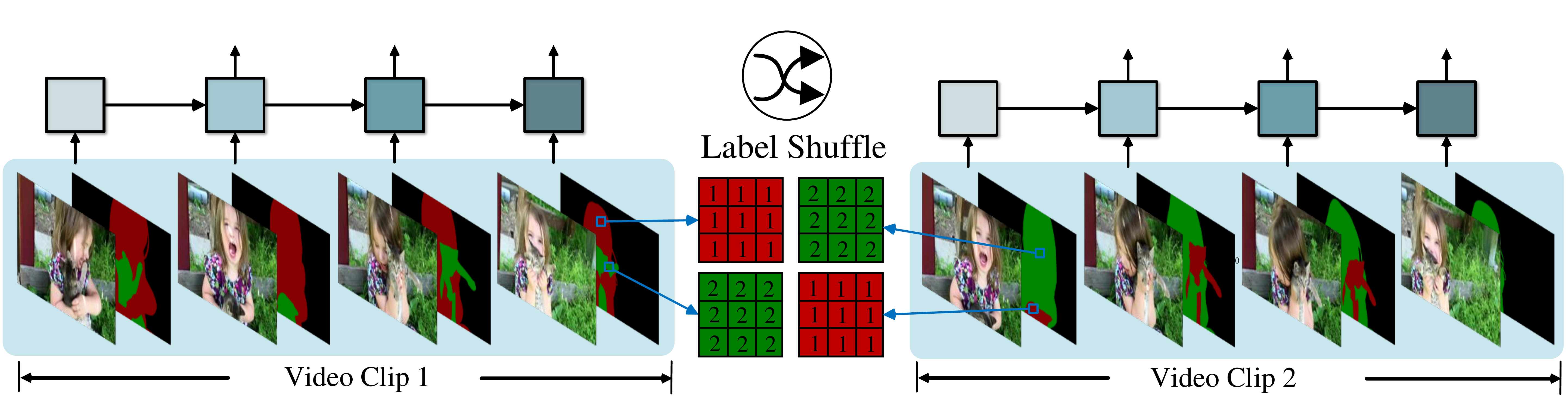}
\put(-295,79.5){\fontsize{7pt}{7pt}\selectfont  {$\bm{\hat{S}}$}}
\put(-259,79.5){\fontsize{7pt}{7pt}\selectfont  {$\bm{\hat{S}}$}}
\put(-224,79.5){\fontsize{7pt}{7pt}\selectfont  {$\bm{\hat{S}}$}}
\put(-99,79.5){\fontsize{7pt}{7pt}\selectfont  {$\bm{\hat{S}}$}}
\put(-65,79.5){\fontsize{7pt}{7pt}\selectfont  {$\bm{\hat{S}}$}}
\put(-30,79.5){\fontsize{7pt}{7pt}\selectfont  {$\bm{\hat{S}}$}}
\put(-330,62){\fontsize{7pt}{7pt}\selectfont  {$\mathcal{G}$}}
\put(-295,62){\fontsize{7pt}{7pt}\selectfont  {$\mathcal{G}$}}
\put(-259,62){\fontsize{7pt}{7pt}\selectfont  {$\mathcal{G}$}}
\put(-224,62){\fontsize{7pt}{7pt}\selectfont {$\mathcal{G}$}}
\put(-134,62){\fontsize{7pt}{7pt}\selectfont  {$\mathcal{G}$}}
\put(-99,62){\fontsize{7pt}{7pt}\selectfont  {$\mathcal{G}$}}
\put(-65,62){\fontsize{7pt}{7pt}\selectfont  {$\mathcal{G}$}}
\put(-30,62){\fontsize{7pt}{7pt}\selectfont  {$\mathcal{G}$}}
	\caption{From video clip to video clip, the instances with associated labels are shuffled. $\mathcal{G}$ denotes the graph memory network and $\bm{\hat{S}}$ is the output prediction.}
	\label{fig:shuffle}
\end{figure*}

\noindent\textbf{Training.} For O-VOS, we train our model following the ``recurrence training'' procedure\!~\cite{wug2018fast,finn2017model}. Each training pass is formed by sampling a support set to build the graph memory and a relevant query set. The core heart of recurrence training is to mimic the inference procedure\!~\cite{behl2018meta}. For each video, we sample $N\!+\!1$ frames to build a support set (first $N$ frames) and a query set (last frame).  Thus, the $N$ support frames can be represented by a $N$-node memory graph. 
We apply a cross entropy loss for supervising training. 

To prevent the graph memory from only memorizing the relation between the instance and the one-hot vector label, we employ the label shuffling strategy\!~\cite{santoro2016meta}. As shown in Fig.\!~\ref{fig:shuffle}, every time we run a segmentation episode, we shuffle the one-hot instance labels\!~\cite{voigtlaender2019feelvos}, meaning sometimes the label of specific instance becomes 2 instead of 1 and vice versa. This encourages the segmentation  network to learn to distinguish the specific instance in current frame by considering the current training samples rather than memorizing specific relation between the target and the given label. See \S\ref{sec:ablation} for detailed experiments for the efficacy of our label shuffling strategy.

To further boost performance, we extend the training set with synthetic videos~\cite{wug2018fast,voigtlaender2019feelvos,Oh_2019_ICCV}. Specifically, for a static image, the video generation technique\!~\cite{perazzi2017learning} is adopted to obtain simulated video clips, through different transformation operations (\eg, rotation, scaling, translation and sheering). The static images come from existing image segmentation datasets.  After pre-training on the synthetic videos, we use the real video data for finetuning.

For Z-VOS, we follow a similar training protocol as O-VOS, but the input modality only has RGB data. We do not use the label shuffling strategy, as we focus on an object-level Z-VOS setting (\ie, do not discriminate different object instances). More training details for Z-VOS and O-VOS can be found in \S\ref{sec:implementation}.

\noindent\textbf{Inference.} 
After training, we directly apply the learned network to unseen test videos without online finetuning. For O-VOS, we process each testing video in a sequential manner. For the first $N$ frames, we compute the memory summarization (Eq.\!~(\ref{eq:retrival})) directly and write these frames into the memory.  From $(N\!+\!1)^{th\!\!}$ frame, after segmentation, we would use this frame to update the graph memory.     Considering the first frame  and its annotation always provide the most reliable information, we re-initialize the node which stores the information about the first frame. Therefore, we use the first annotated frame, last segmented frame and $N\!-\!2$ frames sampled from previous segmented frames, as well as their pre-defined or segmented masks to build the memory. For multiple-instances cases, we run our model for each instance independently and obtain a soft-max probability mask for each one.  Considering the underlying probability overlap between different instances, we combine these results together with a \textit{soft-aggregation} strategy\!~\cite{wug2018fast}. Our network achieves fast segmentation speed of 0.2 s per frame.

For Z-VOS, we randomly sample $N$  frames from the same video to build the graph memory, then we process each frame based on the constructed memory. Considering the global information is more important than local information for handling underlying object occlusions and camera movements, we process each frame independently by re-initializing the graph memory with globally sampled frames. Following common practice\!~\cite{DBLP:conf/iccv/TokmakovAS17,Song_2018_ECCV,cheng2017segflow},  we employ CRF\!~\cite{krahenbuhl2011efficient} binarization and the whole processing speed is about 0.3 s per frame.
\subsection{Implementation Details}
\label{sec:implementation}
 During pre-training, we randomly crop $384\!\times\!384$ patches from static image samples, and the video clip length is $N\!+\!1\!=\!4$. During the main training, we crop $384\!\times\!640$ patches from real training videos. We randomly sample four temporal ordered frames from the same video with a maximum skip of 25 frames to build a training clip. Data augmentation techniques, like rotation, flip and saturation, are adopted to increase the data diversity.
For O-VOS, we select a saliency dataset, MSRA10K\!~\cite{cheng2015global}, and semantic segmentation dataset, COCO\!~\cite{lin2014microsoft},  to synthesize videos.  We use all the training videos in DAVIS$_{17}$\!~\cite{pont20172017} and Youtube-VOS\!~\cite{xu2018youtube-eccv} for the main training.
For Z-VOS, we use two image saliency datasets, MSRA10K\!~\cite{cheng2015global} and DUT\!~\cite{DBLP:conf/cvpr/YangZLRY13}, to generate simulated videos. These two datasets are selected following conventions\!~\cite{DBLP:conf/iccv/TokmakovAS17,lu2019see} for fair comparison. After that, we take advantage of the training set of DAVIS$_{16}$\!~\cite{perazzi2016benchmark} to finetune the network.

 Our model is implemented on \textit{PyTorch} and trained on four  NVIDIA Tesla V100 GPUs with 32 GB memory per card. The batch size is set to 16. We optimize the loss function with Adam optimizer using ``poly'' learning schedule, with the base learning rate of $1e$-$5$ and power of $1.0$. The pre-training stage takes about 24 hours and the main training stage takes about 16 hours for O-VOS. 

\section{Experiments}
\label{sec:exp}
To verify the effectiveness and generic applicability of the proposed method, we perform experiments on different VOS settings. In concrete, we first evaluate our model on two O-VOS datasets (\S\ref{sec:exp1}) and then test it on two Z-VOS datasets (\S\ref{sec:exp2}). Finally, in \S\ref{sec:ablation}, agnostic experiments are conducted for in-depth analysis.
\subsection{Performance for O-VOS}\label{sec:exp1}
\noindent\textbf{Datasets.} Experiments are conducted on two well-known O-VOS benchmarks: DAVIS$_{17}$\!~\cite{pont20172017} and Youtube-VOS\!~\cite{xu2018youtube-eccv}. DAVIS$_{17}$ comprises 60 videos for training and 30 videos for validation. Each video contains one or multiple annotated   object instances. Youtube-VOS is a large-scale dataset which is split into  a \texttt{train} set (3,471 videos) and a \texttt{val} set (474 videos).  The validation set is further divided into \texttt{seen} subset which has the same categories (65) as the \texttt{train} set and \texttt{unseen} subset with 26 unseen categories.

\noindent\textbf{Evaluation Criteria.} Following the standard evaluation protocol of DAVIS$_{17}$,  the mean region similarity $\mathcal{J}$ and contour accuracy $\mathcal{F}$ are reported. For Youtube-VOS, these two metrics are separately computed for the \texttt{seen} and \texttt{unseen} sets.

\begin{table*}[t]
	\centering
	\caption{{\textbf{Evaluation of O-VOS on DAVIS$_{17}$} \texttt{val} set (\S\ref{sec:exp1})}, with region similarity $\mathcal{J}$, boundary accuracy $\mathcal{F}$ and average of $\mathcal{J}\&\mathcal{F}$. Speed is also reported.} 
		\resizebox{0.99\textwidth}{!}{
			\setlength\tabcolsep{7pt}
			\renewcommand\arraystretch{1.1}
			\begin{tabular}{lc||cccccccc}
				\hline\thickhline
				\rowcolor{mygray}
				&&OSMN &SIMMASK&FAVOS &RVOS & OSVOS &AGAME&OnAVOS &RGMP\\
\specialrule{0em}{-0.5pt}{-1pt}
\rowcolor{mygray}
				\multicolumn{2}{c||}{\multirow{-2}{*}{Method}}&\cite{yang2018efficient} &\cite{wang2019fast}&\cite{cheng2018fast} &\cite{ventura2019rvos} &\cite{Cae+17} &\cite{johnander2019generative}&\cite{voigtlaender2017online} & \cite{wug2018fast}\\
\hline
\hline
				$\mathcal{J}$\&$\mathcal{F}$& Mean~$\uparrow$& 54.8& 65.4 &58.2&60.6& 60.3& 71.0&65.4 &66.7 \\ \hline
				\multirow{3}{*}{$\mathcal{J}$} & Mean~$\uparrow$ & 52.5&54.3&54.6&57.5& 56.6& 68.5& 61.6&64.8\\
				&Recall$\uparrow$& 60.9& 62.8& 61.1& 65.2& 63.8& 78.4& 67.4& 74.1\\
				&Decay$\downarrow$& 21.5& 19.3& 14.1& 24.9& 26.1& 14.0& 27.9&18.9 \\ \hline
				\multirow{3}{*}{$\mathcal{F}$} & Mean~$\uparrow$ & 57.1& 58.5& 61.8& 63.6& 63.9& 73.6& 69.1& 68.6\\
				&Recall$\uparrow$ & 66.1& 67.5& 72.3&73.2& 73.8& 83.4& 75.4& 77.7\\
				&Decay$\downarrow$& 24.3& 21.0& 18.0& 28.2& 27.0& 15.8&26.6& 19.6\\
\hline
\multicolumn{2}{c||}{Times (s)}& 0.13 & 0.028& 1.8& 1.8&7.0&0.07& 13&0.13\\
\hline\thickhline
\rowcolor{mygray}
				&&OSVOS-S &RANet &FEELVOS &CINM&PReMVO&DMMNet&STM& \\
\specialrule{0em}{-0.5pt}{-1pt}
\rowcolor{mygray}
\multicolumn{2}{c||}{\multirow{-2}{*}{Method}} &\cite{maninis2018video} &\cite{wang2019ranet} &\cite{voigtlaender2019feelvos} &\cite{bao2018cnn}&\cite{luiten2018premvos}&\cite{Zeng_2019_ICCV}&\cite{Oh_2019_ICCV}&\multirow{-2}{*}{\textbf{Ours}}\\
\hline
\hline
				$\mathcal{J}$\&$\mathcal{F}$& Mean~$\uparrow$& 68.0&65.7& 71.5& 70.6&77.8& 70.7 &81.8& \textbf{82.8} \\
\hline
				\multirow{3}{*}{$\mathcal{J}$} & Mean~$\uparrow$ & 64.7& 63.2& 69.1& 67.2& 73.9&68.1& 79.2& \textbf{80.2}\\
				&Recall$\uparrow$& 74.2& 73.7& 79.1& 74.5& 83.1&77.3&88.7& \textbf{90.1} \\
				&Decay$\downarrow$& 15.1& 18.6& 17.5& 24.6& 16.2&16.8&\textbf{8.0}&6.0 \\ \hline
				\multirow{3}{*}{$\mathcal{F}$} & Mean~$\uparrow$ & 71.3& 68.2& 74.0& 74.0& 81.8& 73.3&84.3&\textbf{85.2}  \\
				&Recall$\uparrow$ & 80.7& 78.8& 83.8& 81.6& 88.9&82.7&91.8& \textbf{93.3} \\
				&Decay$\downarrow$& 18.5& 19.7& 20.1& 26.2& 19.5&23.5&\textbf{10.5}& 8.4  \\
				\hline
				\multicolumn{2}{c||}{Times (s)}& 4.5 & 0.13& 0.5& 38&70&2.7& 0.18&0.2\\
				\hline
			\end{tabular}
		}
	\label{OVOSdavis17}	
\end{table*}

\noindent\textbf{Quantitative Results.}
The performance of our network on DAVIS$_{17}$ is shown in {Table\!~\ref{OVOSdavis17}} with both online learning and offline approaches. Overall, our model outperforms all the contemporary methods and sets a new state-of-the-art in terms of mean $\mathcal{J}\&\mathcal{F}$ (82.8\%), mean $\mathcal{J}$ (80.2\%) and mean $\mathcal{F}$ (85.2\%). Notably, our
method obtains a much higher score for both region similarity and contour accuracy compared to several representative online learning methods: OSVOS\!~\cite{Cae+17}, OnAVOS\!~\cite{voigtlaender2017online}, AGAME\!~\cite{johnander2019generative} and DMMNet\!~\cite{Zeng_2019_ICCV}.   Furthermore, we report the segmentation speed and memory comparison by averaging the inference times for all instances. We observe that most segmentation-by-detection methods (\eg, OSVOS\!~\cite{Cae+17}) consume small GPU memory but need a long time for first frame finetuning and online learning. Meanwhile, most matching based methods (\eg, AGAME\!~\cite{johnander2019generative}, FEELVOS\!~\cite{voigtlaender2019feelvos}, and RGMP\!~\cite{wug2018fast}) achieve fast inference yet suffer from heavy memory cost. However, our method achieves better performance with fast speed and acceptable memory usage.

Moreover, we report the segmentation results on Youtube-VOS in {Table\!~\ref{OVOS-youtube}}. Our approach obtains a final score of 80.2\%, significantly outperforming state-of-the-arts. Compared to memory-based method S2S\!~\cite{xu2018youtube-eccv}, our model achieves much higher performance (\ie, 80.2\% \textit{vs} 64.4\%), which verifies the effectiveness of our external graph memory design.  Moreover, our method performs favorably on both \textit{seen} and \textit{unseen} categories.  Overall, our method achieves huge performance promotion over time-consuming online learning base methods without invoking online finetuning.  This demonstrates the efficacy of our core idea of formulating O-VOS as a procedure of learning to update segmentation network.

\begin{table}[t!]
	\centering
	\caption{{\textbf{Evaluation of O-VOS on Youtube-VOS} \texttt{val} set (\S\ref{sec:exp1}), with region similarity $\mathcal{J}$ and boundary accuracy $\mathcal{F}$. ``Overall": averaged over the four metrics}.\!\!\!
	}
	\resizebox{0.99\textwidth}{!}{
		\setlength\tabcolsep{2pt}
		\renewcommand\arraystretch{1.3}
		\begin{tabular}{lc||cccccccccccccc}
			\hline\thickhline
				\rowcolor{mygray}
		    &{}& MSK & OSMN & RGMP& OnAVOS &RVOS& OSVOS& S2S& AGAME&PreMVOS&DMMNet&STM &\\
		   \rowcolor{mygray}
\specialrule{0em}{-0.5pt}{-3pt}
		   \multicolumn{2}{c||}{\multirow{-2}{*}{Method}}&\!~\cite{perazzi2017learning} & \!~\cite{yang2018efficient}&\!~\cite{wug2018fast}&\!~\cite{voigtlaender2017online}&\!~\cite{ventura2019rvos}&\!~\cite{Cae+17}& \!~\cite{xu2018youtube-eccv}& \!~\cite{johnander2019generative}& \!~\cite{luiten2018premvos}&\!~\cite{Zeng_2019_ICCV}&\!~\cite{Oh_2019_ICCV}&\multirow{-2}{*}{\textbf{Ours}} \\
\specialrule{0em}{-0.5pt}{-0.5pt}
\hline \hline
		\multicolumn{2}{c||}{Overall}& 53.1& 51.2 &53.8& 55.2& 56.8& 58.8& 64.4& 66.1& 66.9& 58.0& 79.4&\textbf{80.2} \\ \hline
			\multirow{2}{*}{\rotatebox{90}{\tabincell{c}{\textit{seen}}}}& Mean$\mathcal{J}$~$\uparrow$  &59.9& 60.0& 59.5& 60.1& 63.6& 59.8& 71.0& 67.8& 71.4& 60.3&79.7& \textbf{80.7} \\
		&	Mean$\mathcal{F}$~$\uparrow$&59.5& 60.1&-& 62.7& 67.2& 60.5&70.0&-&75.9& 63.5& 84.2 & \textbf{85.1} \\ \hline
			\multirow{2}{*}{\rotatebox{90}{\tabincell{c}{\textit{unseen}}}}&Mean$\mathcal{J}$~$\uparrow$	&45.0&40.6& 45.2& 46.6&45.5&54.2&55.5&60.8&56.5&50.6&72.8 &\textbf{74.0}\\
			&Mean$\mathcal{F}$~$\uparrow$&47.9&44.0& -&51.4&51.0&60.7&61.2&-&63.7&57.4&80.9 &\textbf{80.9} \\ \hline
		\end{tabular}
	}
	\label{OVOS-youtube}	
\end{table}

\begin{figure*}[t]
	\centering
	\includegraphics[width=.95\textwidth]{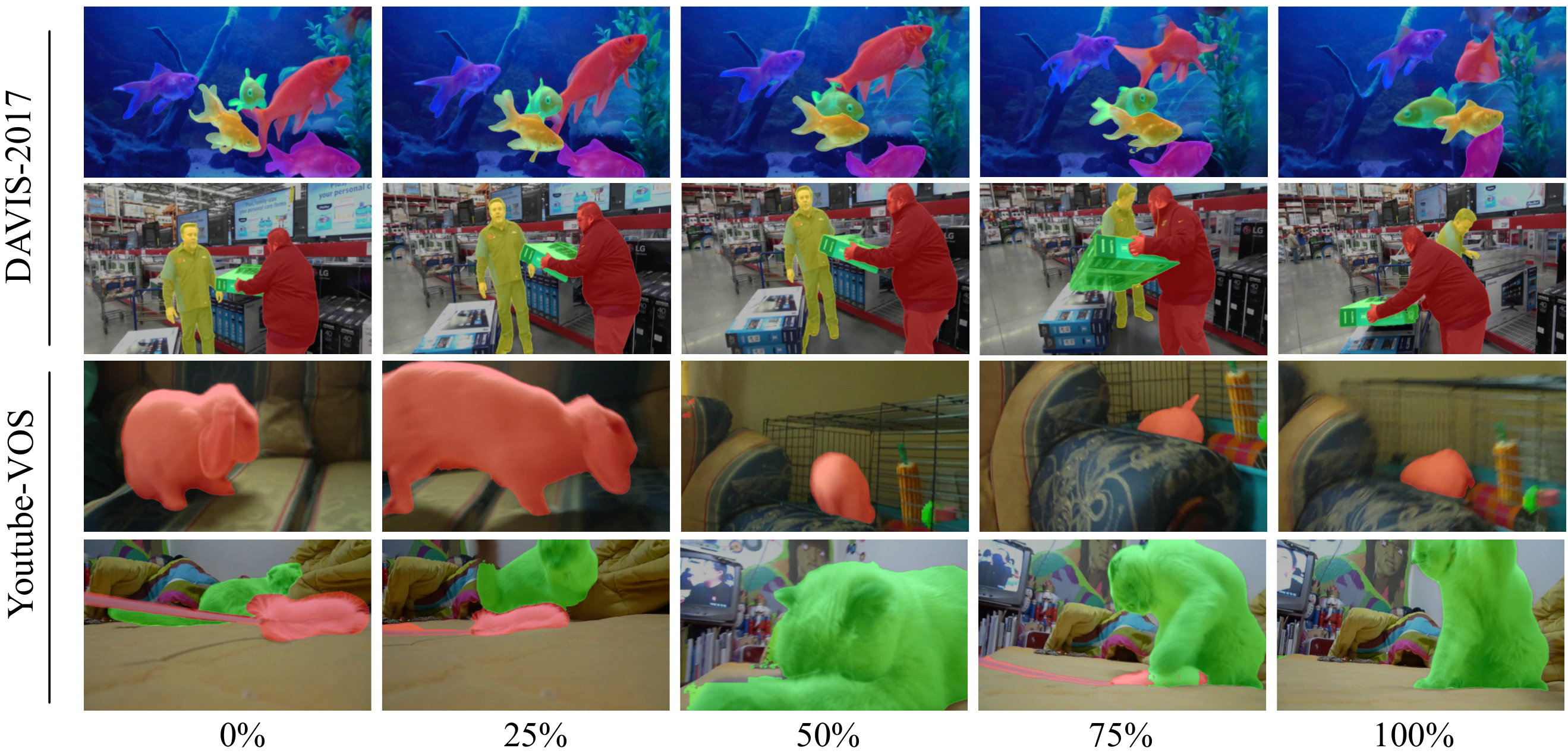}
	\caption{\textbf{Qualitative O-VOS results} on DAVIS$_{17}$ and Youtube-VOS (\S\ref{sec:exp1}).}
	\label{fig:qualitative}
\end{figure*}

\noindent\textbf{Qualitative Results.} In {Fig.\!~\ref{fig:qualitative}}, we show qualitative results of our method at different time steps (uniformly sampled percentage w.r.t. the whole video length) on a few representative videos. Specifically, many instances in the first two DAVIS$_{17}$ videos undergo fast motion and background clutter. However, through the graph memory mechanism, our segmentation network can handle these challenging factors well. The last two Youtube-VOS videos present challenges that the instances suffer occlusion and out-of-view. Once the occlusion ends, our graph memory allows the segmentation network to re-detect the target and segment it successfully.

\subsection{Performance for Z-VOS}\label{sec:exp2}
\noindent\textbf{Datasets.} Experiments are conducted on two challenging datasets: DAVIS$_{16}$\!~\cite{perazzi2016benchmark} and Youtube-Objects\!~\cite{prest2012learning}.  DAVIS$_{16}$ contains 50 videos with 3,455 frames, covering a wide range of challenges, such as fast motion and occlusion. It is split into a \texttt{train} set (30 videos) and a \texttt{val} set (20 videos).
Youtube-Objects has 126 videos belonging to 10 categories and 25,673 frames in total. The \texttt{val} set of DAVIS$_{16}$ and whole Youtube-Objects are used for evaluation.

\begin{table*}[t]
	\centering
	\caption{{\textbf{Evaluation of Z-VOS on DAVIS$_{16}$} \texttt{val} set\!~\cite{perazzi2016benchmark} (\S\ref{sec:exp2})},  with region similarity $\mathcal{J}$, boundary accuracy $\mathcal{F}$ and time
stability $\mathcal{T}$.}  
	\resizebox{0.99\textwidth}{!}{
		\setlength\tabcolsep{7.5pt}
		\renewcommand\arraystretch{1.2}
		\begin{tabular}{lc||ccccccccc}
			\hline\thickhline
			\rowcolor{mygray}
			&&MSG&NLC &CUT &FST &SFL&LMP &FSEG&LVO&UOVOS\\
			\specialrule{0em}{-0.5pt}{-1pt}
\rowcolor{mygray} \multicolumn{2}{c||}{\multirow{-2}{*}{Method}}&\cite{DBLP:conf/iccv/OchsB11}&\cite{DBLP:conf/bmvc/FaktorI14} &\cite{DBLP:conf/iccv/KeuperAB15} &\cite{DBLP:conf/iccv/PapazoglouF13} &\cite{cheng2017segflow}&\cite{DBLP:conf/cvpr/TokmakovAS17} &\cite{jain2017fusionseg}&\cite{DBLP:conf/iccv/TokmakovAS17}&\cite{zhuo2019unsupervised}\\
			\hline
			\hline
			\multirow{3}{*}{$\mathcal{J}$} & Mean~$\uparrow$ & 53.3&55.1&55.2&55.8& 64.7& 70.0& 70.7&75.9&73.9\\
			&Recall$\uparrow$& 61.6& 55.8& 57.5& 64.9& 81.4& 85.0& 83.0& 89.1&88.5\\
			&Decay$\downarrow$& 2.4& 12.6& 2.2& \textbf{0.0}& 6.2& 1.3& 1.5&\textbf{0.0}&0.6 \\ \hline
			\multirow{3}{*}{$\mathcal{F}$} & Mean~$\uparrow$ & 50.8& 52.3& 55.2& 51.1& 66.7& 65.9& 65.3& 72.1&68.0\\
			&Recall$\uparrow$ & 60.0& 61.0& 51.9&51.6& 77.1& 79.2& 73.8& 83.4&80.6\\
			&Decay$\downarrow$& 5.1& 11.4& 3.4& 2.9& 5.1& 2.5&1.8& 1.3&0.7\\
			\hline
			$\mathcal{T}$& Mean~$\downarrow$& 54.8& 65.4 &58.2&60.6& 60.3& 71.0&65.4 &66.7 & 39.0\\ \hline
			\hline\thickhline
			\rowcolor{mygray}
			&  &ARP &PDB &MotAdapt&LSMO&AGS&COSNet&AGNN&AnDiff& \\
			\specialrule{0em}{-0.5pt}{-1pt}
\rowcolor{mygray}
			\multicolumn{2}{c||}{\multirow{-2}{*}{Method}}  &\cite{DBLP:conf/cvpr/KohK17} &\cite{Song_2018_ECCV} &\cite{siam2018video}&\cite{Tokmakov2019}&\cite{wang2019learning}&\cite{lu2019see}&\cite{wang2019zero}&\cite{yang2019anchor}&\multirow{-2}{*}{\textbf{Ours}}\\
			\hline
			\hline
			\multirow{3}{*}{$\mathcal{J}$} & Mean~$\uparrow$ & 76.2& 77.2& 77.2& 78.2& 79.7&80.5& 80.7&81.7& \textbf{82.5}\\
			&Recall$\uparrow$& 89.1& 91.1& 93.1& 87.8& 91.1&93.1&94.0&90.9& \textbf{94.3} \\
			&Decay$\downarrow$& 7.0& 0.9& 5.0& 4.1& 1.9&4.4&0.03&2.2 & 4.2\\ \hline
			\multirow{3}{*}{$\mathcal{F}$} & Mean~$\uparrow$ & 65.3& 72.1& 70.6& 74.5& 77.4& 79.4&79.1&80.5& \textbf{81.2} \\
			&Recall$\uparrow$ & 83.4& 83.5& 84.4& 84.7& 85.8&89.5&\textbf{90.5}&85.1 & 90.3\\
			&Decay$\downarrow$& 7.9& -0.2& 3.3& 3.5& 0.0&5.0&0.03&0.6&5.6 \\\hline
			$\mathcal{T}$& Mean~$\downarrow$& 39.3&29.1& 27.9& 21.2&26.7& \textbf{18.4}&33.7&21.4&19.8 \\
			\hline
			
		\end{tabular}
	}
	\label{ZVOSdavis16}	
\end{table*}

\noindent\textbf{Evaluation Criteria.}  We follow the official
evaluation protocols\!~\cite{perazzi2016benchmark,DBLP:conf/iccv/PapazoglouF13} and report the region similarity $\mathcal{J}$, boundary accuracy $\mathcal{F}$ and time stability $\mathcal{T}$ for DAVIS$_{16}$. Youtube-Objects is evaluated in terms of region similarity $\mathcal{J}$.

\noindent\textbf{Quantitative Results.} For DAVIS$_{16}$\!~\cite{perazzi2016benchmark}, we compare our method with 17 state-of-the-arts  from DAVIS$_{16}$ benchmark\footnote{{\url{https://davischallenge.org/davis2016/soa_compare.html}}} in {{Table\!~\ref{ZVOSdavis16}}}. Our method outperforms other competitors across most metrics. In particular, compared with recent matching-based methods: COSNet\!~\cite{lu2019see}, AGNN\!~\cite{wang2019zero} and AnDiff\!~\cite{zhuo2019unsupervised}, our method achieves an average $\mathcal{J}$ score of 82.5\% which is 0.8\% better than the second best method, AnDiff\!~\cite{zhuo2019unsupervised} despite the fact that it utilizes more training samples than ours. Compared with COSNet\!~\cite{lu2019see}, our method achieves significant performance promotion of 2.0\% and 1.8\% in terms of mean $\mathcal{J}$ and mean $\mathcal{F}$, respectively. Notably, our method outperforms online learning based methods (\ie, SFL\!~\cite{cheng2017segflow}, UVOS\!~\cite{zhuo2019unsupervised} and LSMO\!~\cite{Tokmakov2019}) by a large margin.

We further report results on Youtube-Objects in {Table~\ref{ZVOS-youtube}} with detailed category-wise performance as well as the final average $\mathcal{J}$ score. As seen, our method surpasses other competitors significantly (reaching 71.4\% mean $\mathcal{J}$), especially compared with recent matching based methods\!~\cite{lu2019see,wang2019zero}. Overall, our model consistently yields promising results over different datasets, which clearly illustrates its superior performance and powerful generalizability.

\begin{figure*}[t!]
	\centering
	\includegraphics[width=.95\textwidth]{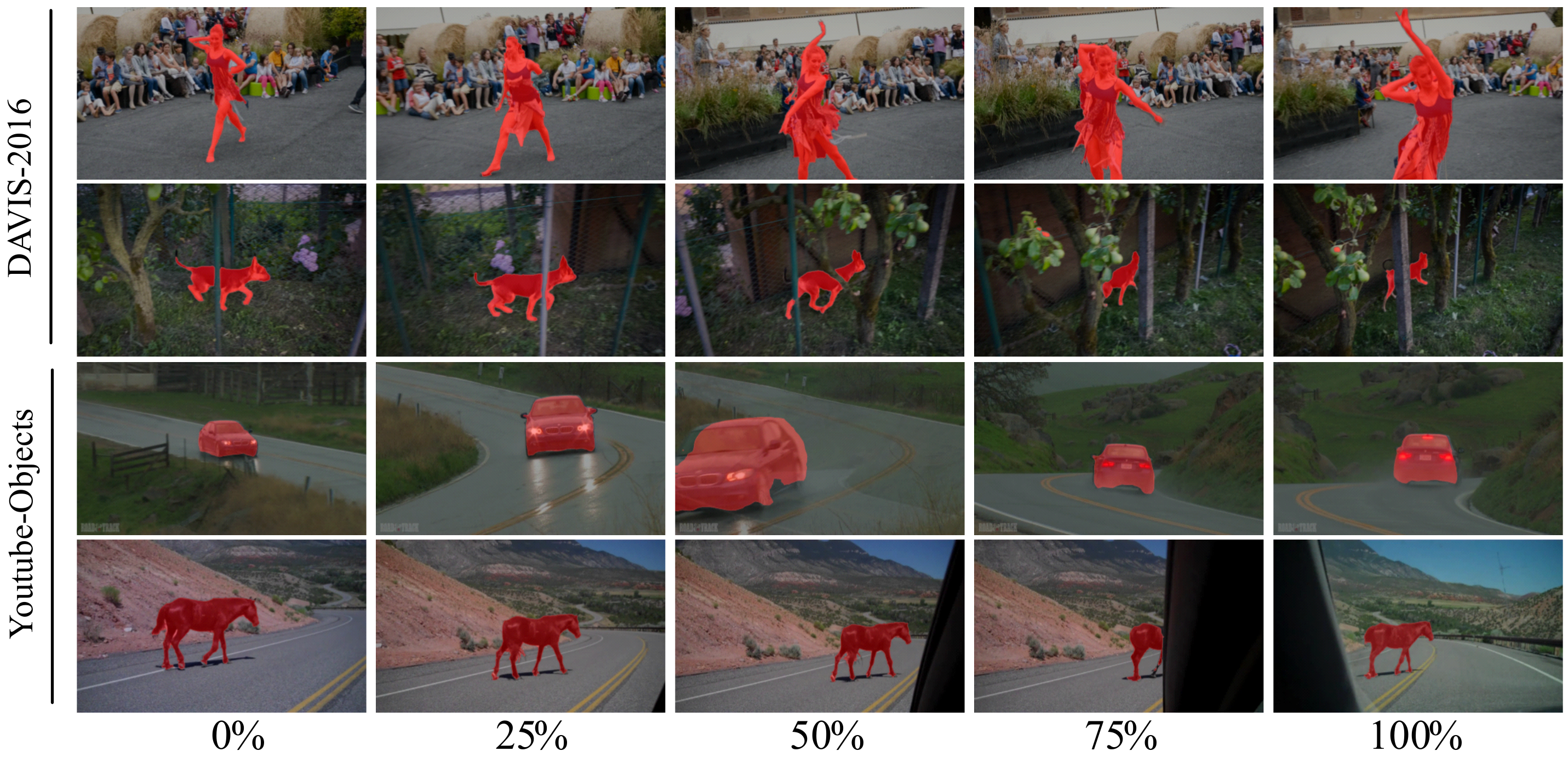}
\caption{\textbf{Qualitative Z-VOS results} on DAVIS$_{16}$ and Youtube-Objects (\S\ref{sec:exp2}).}	
	\label{fig:qualitative1}
\end{figure*}

\noindent\textbf{Quantitative Results.} {Fig.\!~\ref{fig:qualitative1}} depicts some representative visual results on
DAVIS$_{16}$ and Youtube-Objects. As can be observed,  our method handles well these challenging scenes, typically with fast motion, partial occlusion and view changes, even without explicit foreground object indication.

\begin{table}[t]
	\centering
\caption{{\textbf{Evaluation of Z-VOS on Youtube-Objects}\!~\cite{prest2012learning} (\S\ref{sec:exp2}). ``Mean $\mathcal{J}$$\uparrow$" denotes the results averaged over all the categories.}
	}
	\resizebox{0.99\textwidth}{!}{
		\setlength\tabcolsep{2pt}
		\renewcommand\arraystretch{1.2}
		\begin{tabular}{l||cccccccccccccc}
			\hline\thickhline
			\rowcolor{mygray}
			&LTV& FST & COSEG & ARP& LVO &PDB& FSEG& SFL& MotAdapt&LSMO&AGS&COSNet&AGNN &\\
			\rowcolor{mygray}
			\specialrule{0em}{-0.5pt}{-3pt}
			\multirow{-2}{*}{Method}&\!~\cite{DBLP:journals/pami/OchsMB14}&\!~\cite{DBLP:conf/iccv/PapazoglouF13} & \!~\cite{tsai2016semantic}&\!~\cite{DBLP:conf/cvpr/KohK17}&\!~\cite{DBLP:conf/iccv/TokmakovAS17}&\!~\cite{Song_2018_ECCV}& \!~\cite{jain2017fusionseg}& \!~\cite{cheng2017segflow}&\!~\cite{siam2018video}& \!~\cite{Tokmakov2019}&\!~\cite{wang2019learning}&\!~\cite{lu2019see}&\!~\cite{wang2019zero}&\multirow{-2}{*}{\textbf{Ours}} \\
			\hline \hline
			Airplane (6) &13.7& 70.9 &69.3 &73.6 &86.2 &78.0&{81.7}&65.6 &77.2& 60.5&{{87.7}}& 81.1&81.1&{86.1}\\		
			Bird (6) &12.2 &70.6 &76.0 &56.1 &{{81.0}} & {{80.0}}& 63.8 & 65.4& 42.2& 59.3& 76.7&75.7&75.9&75.7\\
			Boat (15) & 10.8 & 42.5& 53.5 &57.8 &68.5& 58.9& {{72.3}}& 59.9 &49.3& 62.1&{72.2}&71.3&70.7&68.6\\
			Car (7) & 23.7&65.2 &70.4 &33.9 &69.3& 76.5&74.9&64.0 &68.6&72.3&{78.6}&77.6&78.1&{{82.4}}\\
			Cat (16) & 18.6&52.1 &66.8 &30.5 &58.8 &63.0&{68.4}&58.9& 46.3 &66.3&{{69.2}}&66.5&67.9&65.9\\
			Cow (20)&16.3&44.5 &49.0 & 41.8&68.5&64.1&68.0 &51.1& 64.2 &67.9&64.6 &{69.8}&69.7&{{70.5}}\\
			Dog (27)&18.2 &65.3 &47.5 &36.8&61.7 &70.1&69.4& 54.1 & 66.1&70.0&73.3&{76.8}&77.4&{{77.1}}\\
			Horse (14)& 11.5& 53.5 &55.7 &44.3 &53.9 &{67.6}&60.4&64.8&64.8& 65.4 &64.4&{67.4}&67.3&{{72.2}}\\
			Motorbike (10)& 10.6 &44.2 &39.5 & 48.9&60.8&58.3&62.7 &52.6 &44.6&55.5&62.1&{{67.7}}&68.3&{63.8}\\
			Train (5) &19.6 &29.6 &53.4 &39.2 &{{66.3}}&35.2&{62.2}& 34.0 & 42.3&38.0&48.2&46.8&47.8&{47.8}\\\hline
			\textit{Mean} $\mathcal{J}$$\uparrow$ & 15.5&53.8 &58.1 &46.2 &67.5&65.4&{68.4}& 57.0& 58.1&64.3& {69.7}&{70.5}&70.8&{\textbf{71.4}}\\ 	\hline
		\end{tabular}
	}
	\label{ZVOS-youtube}
\end{table}

\subsection{Diagnostic Experiments}
\label{sec:ablation}
In this section, we analyze the contribution of different model components to the final performance.
Specifically, we take O-VOS and Z-VOS as exemplar and evaluate all ablated versions on DAVIS$_{17}$\!~\cite{pont20172017} and DAVIS$_{16}$\!~\cite{perazzi2016benchmark}, respectively.  The experimental results are evaluated by mean $\mathcal{J}$ and mean $\mathcal{F}$.
For each ablated version, we retrain the model from scratch using the same protocol. From the whole results comparison in {Table~\ref{table:abl}}, we can draw several essential conclusions.

\noindent\textbf{Graph Memory Network:} First, with the proposed graph memory network, our method yields significant  performance improvements (+6.5\%,+7.5\% and +9.3\%, +8.7\% in terms of mean $\mathcal{J}$ and mean $\mathcal{F}$) than the backbone over different VOS settings. This supports our view that the graph memory network with differential controllers can learn to update the segmentation network effectively.

\begin{table}[t!]
	\centering
	\caption{{\textbf{Ablation study} of our graph memory network (\S\ref{sec:ablation}). }
	}
	\resizebox{0.95\textwidth}{!}{
		\setlength\tabcolsep{4pt}
		\renewcommand\arraystretch{1.06}
		\begin{tabular}{c|c||cc|cc}
			\hline\thickhline
			\rowcolor{mygray}
			&&\multicolumn{2}{c|}{One-shot VOS}&\multicolumn{2}{c}{ Zero-shot VOS}\\\specialrule{0em}{-0.5pt}{-3pt}
			\rowcolor{mygray}
			&&\multicolumn{2}{c|}{DAVIS$_{17}$\!~\cite{pont20172017}} &\multicolumn{2}{c}{DAVIS$_{16}$\!~\cite{perazzi2016benchmark}} \\ \cline{3-6}
			\rowcolor{mygray}
			\multirow{-3}{*}{Aspect}&\multirow{-3}{*}{Method} &Mean $\mathcal{J}$$\uparrow$ &Mean $\mathcal{F}$$\uparrow$&Mean $\mathcal{J}$$\uparrow$ &Mean $\mathcal{F}$$\uparrow$  \\ \hline \hline
			\specialrule{0em}{-0.5pt}{-1.5pt}
			\textbf{Full model}& graph memory (3 nodes, 3 episodes)& 80.0& 85.9 &82.5& 81.2\\ \hline \hline
			Backbone& direct infer. \textit{w/o} graph memory&73.5& 78.4&73.2& 72.5\\\hline
			\multirow{3}{*}{\tabincell{c}{Graph \\Structure}}& 2 nodes & 76.0 & 81.4& 78.5& 76.8 \\
			&4 nodes&79.5&84.6&82.5&81.2\\
			&5 nodes &80.0& 85.9 &82.5&81.2 \\\hline
			\multirow{3}{*}{\tabincell{c}{State \\ Updating}}& $K=0$ & 78.1 & 82.2& 81.2& 79.7 \\
			& $K=1$& 78.9& 83.3& 81.6&80.3\\
			& $K=2$&79.3&84.8&82.0&80.8\\
			& $K=4$& 80.0&85.9&82.5&81.2\\ \hline
			Training  & \textit{w/o} label shuffling & 78.5& 82.7 &-&- \\
			\hline
		\end{tabular}
	}
	\label{table:abl}	
\end{table}

\noindent\textbf{Memory Size:}
We next investigate the influence of memory size on the final performance (1$^{st\!}$ and 3$^{rd}$-5$^{th\!}$ rows). We find only a 3-node memory is enough for achieving good performance, further verifying the efficacy of our memory design.

\noindent\textbf{Iterative Memory Reasoning:}
It is also of interest to assess the impact of our iterative memory updating strategy. When $K\!=\!0$, it means no update for graph memory network, therefore the state of the network is fixed without online learning. In this case, the results deteriorate significantly. 
We further observe that more steps can boost the performance (1$\rightarrow$3) and when the step is increased to certain extent ($K\!=\!4$), the performance remains almost unchanged.

\noindent\textbf{Label Shuffling:}
Finally we study the effect of our label shuffling strategy. Comparing results on the first and last rows, we can easily observe that shuffling instance labels during network training indeed promotes O-VOS performance.

\section{Conclusion}
This paper integrates  a novel graph memory mechanism to efficiently adapt the segmentation network to specific videos without catastrophic inference/finetune. Through episodic reasoning the memory graph, the proposed model is capable of generating video-specific memory summarization which benefits the final segmentation  prediction significantly.   Meanwhile, the online model updating can be implemented via learnable memory controllers. Our method is effective and principle, which can be easily extended to Z-VOS setting. Extensive experimental results on four challenging datasets demonstrate its promising performance.


\clearpage
%
%
{\small
\bibliographystyle{splncs04}
\bibliography{egbib}

\begin{thebibliography}{10}
\providecommand{\url}[1]{\texttt{#1}}
\providecommand{\urlprefix}{URL }
\providecommand{\doi}[1]{https://doi.org/#1}

\bibitem{DBLP:conf/cvpr/BadrinarayananGC10}
Badrinarayanan, V., Galasso, F., Cipolla, R.: Label propagation in video
  sequences. In: CVPR (2010)

\bibitem{ballas2015delving}
Ballas, N., Yao, L., Pal, C., Courville, A.: Delving deeper into convolutional
  networks for learning video representations. In: ICLR (2016)

\bibitem{bao2018cnn}
Bao, L., Wu, B., Liu, W.: Cnn in mrf: Video object segmentation via inference
  in a cnn-based higher-order spatio-temporal mrf. In: CVPR (2018)

\bibitem{behl2018meta}
Behl, H.S., Najafi, M., Arnab, A., Torr, P.H.: Meta learning deep visual words
  for fast video object segmentation. NeurIPS Workshop  (2019)

\bibitem{Cae+17}
Caelles, S., Maninis, K.K., Pont-Tuset, J., Leal-Taix\'e, L., Cremers, D., {Van
  Gool}, L.: One-shot video object segmentation. In: CVPR (2017)

\bibitem{Chen_2018_CVPR}
Chen, Y., Pont-Tuset, J., Montes, A., Van~Gool, L.: Blazingly fast video object
  segmentation with pixel-wise metric learning. In: CVPR (2018)

\bibitem{cheng2018fast}
Cheng, J., Tsai, Y.H., Hung, W.C., Wang, S., Yang, M.H.: Fast and accurate
  online video object segmentation via tracking parts. In: CVPR (2018)

\bibitem{cheng2017segflow}
Cheng, J., Tsai, Y.H., Wang, S., Yang, M.H.: Segflow: Joint learning for video
  object segmentation and optical flow. In: ICCV (2017)

\bibitem{cheng2015global}
Cheng, M.M., Mitra, N.J., Huang, X., Torr, P.H., Hu, S.M.: Global contrast
  based salient region detection. {IEEE} TPAMI  \textbf{37}(3),  569--582
  (2015)

\bibitem{ci2018video}
Ci, H., Wang, C., Wang, Y.: Video object segmentation by learning
  location-sensitive embeddings. In: ECCV (2018)

\bibitem{Duarte_2019_ICCV}
Duarte, K., Rawat, Y.S., Shah, M.: Capsulevos: Semi-supervised video object
  segmentation using capsule routing. In: ICCV (2019)

\bibitem{DBLP:conf/bmvc/FaktorI14}
Faktor, A., Irani, M.: Video segmentation by non-local consensus voting. In:
  BMVC (2014)

\bibitem{finn2017model}
Finn, C., Abbeel, P., Levine, S.: Model-agnostic meta-learning for fast
  adaptation of deep networks. In: ICML (2017)

\bibitem{he2016deep}
He, K., Zhang, X., Ren, S., Sun, J.: Deep residual learning for image
  recognition. In: CVPR (2016)

\bibitem{Hu_2018_ECCV}
Hu, Y.T., Huang, J.B., Schwing, A.G.: Videomatch: Matching based video object
  segmentation. In: ECCV (2018)

\bibitem{jain2017fusionseg}
Jain, S.D., Xiong, B., Grauman, K.: Fusionseg: Learning to combine motion and
  appearance for fully automatic segmention of generic objects in videos. In:
  CVPR (2017)

\bibitem{johnander2019generative}
Johnander, J., Danelljan, M., Brissman, E., Khan, F.S., Felsberg, M.: A
  generative appearance model for end-to-end video object segmentation. In:
  CVPR (2019)

\bibitem{DBLP:conf/iccv/KeuperAB15}
Keuper, M., Andres, B., Brox, T.: Motion trajectory segmentation via minimum
  cost multicuts. In: ICCV (2015)

\bibitem{DBLP:conf/cvpr/KohK17}
Koh, Y.J., Kim, C.: Primary object segmentation in videos based on region
  augmentation and reduction. In: CVPR (2017)

\bibitem{krahenbuhl2011efficient}
Kr{\"a}henb{\"u}hl, P., Koltun, V.: Efficient inference in fully connected crfs
  with gaussian edge potentials. In: NIPS (2011)

\bibitem{kumar2016ask}
Kumar, A., Irsoy, O., Ondruska, P., Iyyer, M., Bradbury, J., Gulrajani, I.,
  Zhong, V., Paulus, R., Socher, R.: Ask me anything: Dynamic memory networks
  for natural language processing. In: ICML (2016)

\bibitem{lee2011key}
Lee, Y.J., Kim, J., Grauman, K.: Key-segments for video object segmentation.
  In: ICCV (2011)

\bibitem{lin2014microsoft}
Lin, T.Y., Maire, M., Belongie, S., Hays, J., Perona, P., Ramanan, D.,
  Doll{\'a}r, P., Zitnick, C.L.: Microsoft coco: Common objects in context. In:
  ECCV (2014)

\bibitem{lu2019see}
Lu, X., Wang, W., Ma, C., Shen, J., Shao, L., Porikli, F.: See more, know more:
  Unsupervised video object segmentation with co-attention siamese networks.
  In: CVPR (2019)

\bibitem{lu2020learning}
Lu, X., Wang, W., Shen, J., Tai, Y.W., Crandall, D.J., Hoi, S.C.: Learning
  video object segmentation from unlabeled videos. In: CVPR (2020)

\bibitem{luiten2018premvos}
Luiten, J., Voigtlaender, P., Leibe, B.: Premvos: Proposal-generation,
  refinement and merging for video object segmentation. In: ACCV (2018)

\bibitem{maninis2018video}
Maninis, K.K., Caelles, S., Chen, Y., Pont-Tuset, J., Leal-Taix{\'e}, L.,
  Cremers, D., Van~Gool, L.: Video object segmentation without temporal
  information. {IEEE} TPAMI  \textbf{41}(6),  1515--1530 (2018)

\bibitem{miller2016key}
Miller, A., Fisch, A., Dodge, J., Karimi, A.H., Bordes, A., Weston, J.:
  Key-value memory networks for directly reading documents. In: EMNLP (2016)

\bibitem{DBLP:conf/iccv/OchsB11}
Ochs, P., Brox, T.: Object segmentation in video: {A} hierarchical variational
  approach for turning point trajectories into dense regions. In: ICCV (2011)

\bibitem{DBLP:journals/pami/OchsMB14}
Ochs, P., Malik, J., Brox, T.: Segmentation of moving objects by long term
  video analysis. {IEEE} TPAMI  \textbf{36}(6),  1187--1200 (2014)

\bibitem{Oh_2019_ICCV}
Oh, S.W., Lee, J.Y., Xu, N., Kim, S.J.: Video object segmentation using
  space-time memory networks. In: ICCV (2019)

\bibitem{DBLP:conf/iccv/PapazoglouF13}
Papazoglou, A., Ferrari, V.: Fast object segmentation in unconstrained video.
  In: ICCV (2013)

\bibitem{perazzi2017learning}
Perazzi, F., Khoreva, A., Benenson, R., Schiele, B., Sorkine-Hornung, A.:
  Learning video object segmentation from static images. In: CVPR (2017)

\bibitem{perazzi2016benchmark}
Perazzi, F., Pont-Tuset, J., McWilliams, B., Van~Gool, L., Gross, M.,
  Sorkine-Hornung, A.: A benchmark dataset and evaluation methodology for video
  object segmentation. In: CVPR (2016)

\bibitem{perazzi2015fully}
Perazzi, F., Wang, O., Gross, M., Sorkine-Hornung, A.: Fully connected object
  proposals for video segmentation. In: CVPR (2015)

\bibitem{pont20172017}
Pont-Tuset, J., Perazzi, F., Caelles, S., Arbel{\'a}ez, P., Sorkine-Hornung,
  A., Van~Gool, L.: The 2017 davis challenge on video object segmentation.
  arXiv preprint arXiv:1704.00675  (2017)

\bibitem{prest2012learning}
Prest, A., Leistner, C., Civera, J., Schmid, C., Ferrari, V.: Learning object
  class detectors from weakly annotated video. In: CVPR (2012)

\bibitem{rakelly2019metalearning}
Rakelly*, K., Shelhamer*, E., Darrell, T., Efros, A.A., Levine, S.:
  Meta-learning to guide segmentation. In: ICLR (2019)

\bibitem{santoro2016meta}
Santoro, A., Bartunov, S., Botvinick, M., Wierstra, D., Lillicrap, T.:
  Meta-learning with memory-augmented neural networks. In: ICML (2016)

\bibitem{shankar2015video}
Shankar~Nagaraja, N., Schmidt, F.R., Brox, T.: Video segmentation with just a
  few strokes. In: ICCV (2015)

\bibitem{siam2018video}
Siam, M., Jiang, C., Lu, S., Petrich, L., Gamal, M., Elhoseiny, M., Jagersand,
  M.: Video segmentation using teacher-student adaptation in a human robot
  interaction (hri) setting. In: ICRA (2019)

\bibitem{Song_2018_ECCV}
Song, H., Wang, W., Zhao, S., Shen, J., Lam, K.M.: Pyramid dilated deeper
  convlstm for video salient object detection. In: ECCV (2018)

\bibitem{sukhbaatar2015end}
Sukhbaatar, S., Szlam, A., Weston, J., Fergus, R.: End-to-end memory networks.
  In: NIPS (2015)

\bibitem{DBLP:conf/cvpr/TokmakovAS17}
Tokmakov, P., Alahari, K., Schmid, C.: Learning motion patterns in videos. In:
  CVPR (2017)

\bibitem{DBLP:conf/iccv/TokmakovAS17}
Tokmakov, P., Alahari, K., Schmid, C.: Learning video object segmentation with
  visual memory. In: ICCV (2017)

\bibitem{Tokmakov2019}
Tokmakov, P., Schmid, C., Alahari, K.: Learning to segment moving objects. IJCV
   \textbf{127}(3),  282--301 (2019)

\bibitem{tsai2016semantic}
Tsai, Y.H., Zhong, G., Yang, M.H.: Semantic co-segmentation in videos. In: ECCV
  (2016)

\bibitem{ventura2019rvos}
Ventura, C., Bellver, M., Girbau, A., Salvador, A., Marques, F., Giro-i Nieto,
  X.: Rvos: End-to-end recurrent network for video object segmentation. In:
  CVPR (2019)

\bibitem{voigtlaender2019feelvos}
Voigtlaender, P., Chai, Y., Schroff, F., Adam, H., Leibe, B., Chen, L.C.:
  Feelvos: Fast end-to-end embedding learning for video object segmentation.
  In: CVPR (2019)

\bibitem{voigtlaender2017online}
Voigtlaender, P., Leibe, B.: Online adaptation of convolutional neural networks
  for video object segmentation. In: BMVC (2017)

\bibitem{wang2019fast}
Wang, Q., Zhang, L., Bertinetto, L., Hu, W., Torr, P.H.: Fast online object
  tracking and segmentation: A unifying approach. In: CVPR (2019)

\bibitem{wang2019zero}
Wang, W., Lu, X., Shen, J., Crandall, D.J., Shao, L.: Zero-shot video object
  segmentation via attentive graph neural networks. In: ICCV (2019)

\bibitem{wang2018semi}
Wang, W., Shen, J., Porikli, F., Yang, R.: Semi-supervised video object
  segmentation with super-trajectories. {IEEE} TPAMI  \textbf{41}(4),  985--998
  (2018)

\bibitem{wang2017saliency}
Wang, W., Shen, J., Yang, R., Porikli, F.: Saliency-aware video object
  segmentation. IEEE TPAMI  \textbf{40}(1),  20--33 (2017)

\bibitem{wang2019learning}
Wang, W., Song, H., Zhao, S., Shen, J., Zhao, S., Hoi, S.C., Ling, H.: Learning
  unsupervised video object segmentation through visual attention. In: CVPR
  (2019)

\bibitem{wang2019ranet}
Wang, Z., Xu, J., Liu, L., Zhu, F., Shao, L.: Ranet: Ranking attention network
  for fast video object segmentation. In: ICCV (2019)

\bibitem{wen2015jots}
Wen, L., Du, D., Lei, Z., Li, S.Z., Yang, M.H.: Jots: Joint online tracking and
  segmentation. In: CVPR (2015)

\bibitem{weston2014memory}
Weston, J., Chopra, S., Bordes, A.: Memory networks. ICLR  (2015)

\bibitem{wug2018fast}
Wug~Oh, S., Lee, J.Y., Sunkavalli, K., Joo~Kim, S.: Fast video object
  segmentation by reference-guided mask propagation. In: CVPR (2018)

\bibitem{xiao2018monet}
Xiao, H., Feng, J., Lin, G., Liu, Y., Zhang, M.: Monet: Deep motion
  exploitation for video object segmentation. In: CVPR (2018)

\bibitem{xiao2019online}
Xiao, H., Kang, B., Liu, Y., Zhang, M., Feng, J.: Online meta adaptation for
  fast video object segmentation. IEEE transactions on pattern analysis and
  machine intelligence  (2019)

\bibitem{xie2019attentive}
Xie, G.S., Liu, L., Jin, X., Zhu, F., Zhang, Z., Qin, J., Yao, Y., Shao, L.:
  Attentive region embedding network for zero-shot learning. In: CVPR (2019)

\bibitem{xie2020region}
Xie, G.S., Liu, L., Zhu, F., Zhao, F., Zhang, Z., Qin, J., Yao, Y., Shao, L.:
  Region graph embedding network for zero-shot learning. In: ECCV (2020)

\bibitem{xu2018youtube-eccv}
Xu, N., Yang, L., Fan, Y., Yang, J., Yue, D., Liang, Y., Price, B., Cohen, S.,
  Huang, T.: Youtube-vos: Sequence-to-sequence video object segmentation. In:
  ECCV (2018)

\bibitem{DBLP:conf/cvpr/YangZLRY13}
Yang, C., Zhang, L., Lu, H., Ruan, X., Yang, M.: Saliency detection via
  graph-based manifold ranking. In: CVPR (2013)

\bibitem{yang2018efficient}
Yang, L., Wang, Y., Xiong, X., Yang, J., Katsaggelos, A.K.: Efficient video
  object segmentation via network modulation. In: CVPR (2018)

\bibitem{Yang2018}
Yang, T., Chan, A.B.: Learning dynamic memory networks for object tracking. In:
  ECCV (2018)

\bibitem{yang2019anchor}
Yang, Z., Wang, Q., Bertinetto, L., Bai, S., Hu, W., Torr, P.H.: Anchor
  diffusion for unsupervised video object segmentation. In: ICCV (2019)

\bibitem{DBLP:conf/iccv/YoonRKLSK17}
Yoon, J.S., Rameau, F., Kim, J., Lee, S., Shin, S., Kweon, I.S.: Pixel-level
  matching for video object segmentation using convolutional neural networks.
  In: ICCV (2017)

\bibitem{Zeng_2019_ICCV}
Zeng, X., Liao, R., Gu, L., Xiong, Y., Fidler, S., Urtasun, R.: Dmm-net:
  Differentiable mask-matching network for video object segmentation. In: ICCV
  (2019)

\bibitem{zhang2013}
Zhang, D., Javed, O., Shah, M.: Video object segmentation through spatially
  accurate and temporally dense extraction of primary object regions. In: CVPR
  (2013)

\bibitem{zhou2020motion}
Zhou, T., Wang, S., Zhou, Y., Yao, Y., Li, J., Shao, L.: Motion-attentive
  transition for zero-shot video object segmentation. In: AAAI (2020)

\bibitem{zhuo2019unsupervised}
Zhuo, T., Cheng, Z., Zhang, P., Wong, Y., Kankanhalli, M.: Unsupervised online
  video object segmentation with motion property understanding. IEEE TIP
  \textbf{29},  237--249 (2019)

\end{thebibliography}
}
\end{document}